\documentclass[runningheads,a4paper]{llncs}

\usepackage{amsmath}
\usepackage{amssymb}
\usepackage{graphicx}

\usepackage{url}
\newcommand{\keywords}[1]{\par\addvspace\baselineskip
\noindent\keywordname\enspace\ignorespaces#1}

\usepackage[utf8]{inputenc}
\usepackage[english]{babel}
\usepackage{multirow}
\usepackage{booktabs}
\usepackage{microtype}
\begin{document}

\mainmatter  % start of an individual contribution

% first the title is needed
\title{How Many Dissimilarity/Kernel Self Organizing Map Variants Do We Need?}

% a short form should be given in case it is too long for the running head
\titlerunning{How Many Dissimilarity/Kernel SOM Variants are Needed?}

% the name(s) of the author(s) follow(s) next
%
% NB: Chinese authors should write their first names(s) in front of
% their surnames. This ensures that the names appear correctly in
% the running heads and the author index.
%
\author{Fabrice Rossi}
\authorrunning{Fabrice Rossi}
% (feature abused for this document to repeat the title also on left hand pages)

% the affiliations are given next; don't give your e-mail address
% unless you accept that it will be published
\institute{SAMM (EA 4543), Universit\'{e} Paris 1,\\
90, rue de Tolbiac, 75634 Paris Cedex 13, France\\
\email{fabrice.rossi@univ-paris1.fr}}

%
% NB: a more complex sample for affiliations and the mapping to the
% corresponding authors can be found in the file "llncs.dem"
% (search for the string "\mainmatter" where a contribution starts).
% "llncs.dem" accompanies the document class "llncs.cls".
%

\toctitle{How Many Dissimilarity/Kernel Self Organizing Map Variants Do We Need?}
\tocauthor{Fabrice Rossi}
\maketitle

\begin{abstract}
In numerous applicative contexts, data are too rich and too complex to be
represented by numerical vectors. A general approach to extend machine
learning and data mining techniques to such data is to really on a
dissimilarity or on a kernel that measures how different or similar two
objects are. 

This approach has been used to define several variants of the Self Organizing
Map (SOM). This paper reviews those variants in using a common set of notations in
order to outline differences and similarities between them. It discusses the
advantages and drawbacks of the variants, as well as the actual relevance of
the dissimilarity/kernel SOM for practical applications.

\keywords{Self Organizing Map; Dissimilarity data; Pairwise data; Kernel;
  Deterministic annealing}
\end{abstract}

\section{Introduction}
Complex data are frequently too rich and too elaborate to be represented in a
simple tabular form where each object is described via a fixed set of
attributes/variables with numerical and/or nominal values. This is especially
the case for relational data when objects of different categories are
interconnected by relations of different types. For instance online retailers
have interconnected customers and products databases, in which a customer can
buy one or several copies of a product, and can also leave some score and/or
review of said products.

Adapting data mining and machine learning methods to complex data is possible,
but time consuming and complex, both at the theoretical level (e.g.,
consistency of the algorithms is generally proved only in the Euclidean case)
and on a practical point of view (new implementations are needed). Therefore,
it is tempting to build generic methods that use only properties that are
shared by all types of data. 

Two such generic approaches have been used successfully: the dissimilarity
based approach and the kernel based approach
\cite{ShaweTaylorChristianini2004KernelMethods}. Both are based on fairly
generic assumptions: the analyst is given a data set on which either a
dissimilarity or a kernel is defined. A dissimilarity measures how much two
objects differs, while a kernel can be seen as a form a similarity measure, at
least in the correlation sense. Dozens of dissimilarities and kernels have
been proposed over the years, covering many types of complex data (see
e.g. \cite{gartner2004kernels}). Then one needs only to adapt a classical data
mining or machine learning method to the dissimilarity/kernel setting in order
to obtain a fully generic approach. As a dissimilarity can always be
constructed from a kernel, dissimilarity algorithms are probably the more
generic ones. A typical example is the $k$ nearest neighbor method which is
based only on dissimilarities.

We review in this paper variants of the Self Organizing Map (SOM) that have
been proposed following this line of research, that is SOM variants that
operate on dissimilarity/kernel data. We discuss whether those variants are
really usable and helpful in practice. The paper is organized as
follows. Section \ref{sec:general-setting} describes our general setting:
dissimilarity data, kernel data and the Self Organizing Map. Section
\ref{sec:median-som} is dedicated to the oldest dissimilarity variant of the
SOM, the Median SOM, while Section \ref{sec:relational-som} focuses on the
modern variant, the relational SOM. Section \ref{sec:soft-topogr-mapp}
presents a different approach to SOM extensions based on the deterministic
annealing principle. Section \ref{sec:kernel-som} describes kernel based
variants of the SOM. An unifying view is provided in Section
\ref{sec:equiv-betw-som} which shows that the differences between the SOM
variants are mainly explained by the optimization strategy rather than by the
data properties. Finally Section \ref{sec:discussion} gathers our personal
remarks and insights on the dissimilarity/kernel SOM variants. 

\section{General setting}\label{sec:general-setting}
The data set under study comprises $N$ data points $x_1,\ldots,x_N$ from an
abstract space $\mathcal{X}$. We specify below the two options, namely
dissimilarity data and kernel data. We also recall the classical SOM
algorithms. 

\subsection{Dissimilarity data}
In the dissimilarity data setting (a.k.a. the pairwise data setting), it is
assumed that the data are described indirectly by a square $N\times N$ symmetric
matrix $D$ that contains dissimilarities between the data points. The
convention is that $D_{ij}=d(x_i,x_j)$, a non negative real number, is high
when $x_i$ and $x_j$ are different and low when they are similar. Minimal
assumptions on $D$ are symmetry and non negativity of each element. It is also
natural to assume some basic ordering, that is that $D_{ii}\leq D_{ij}$ for
all $i$ and $j$, but this is not used in SOM variants. Some theoretical results
also need $D_{ii}=0$ (e.g. \cite{HammerHasenfuss2010NeuralComputation}), but
again this is not a very strong constraint. Notice that one can be given either
the dissimilarity function $d$ from $\mathcal{X}^2$ to $\mathbb{R}^+$ or
directly the matrix $D$.

\subsection{Kernel data}\label{sec:kernel-data}
In the kernel data setting, one is given a \emph{kernel} function $k$ from
$\mathcal{X}^2$ to $\mathbb{R}$ which satisfies the following properties:
\begin{enumerate}
\item $k$ is symmetric: for all $x$ and $y$ in $\mathcal{X}$, $k(x,y)=k(y,x)$;
\item $k$ is positive definite: for all $m>0$, all
  observation set $(x_1,\ldots,x_m)\in\mathcal{X}^m$ and all coefficient set 
  $(\alpha_1,\ldots,\alpha_m)\in\mathbb{R}^m$,
  $\sum_{i=1}^m\sum_{j=1}^m\alpha_i\alpha_jk(x_i,x_j)\geq 0$. 
\end{enumerate}
The most important aspect of the kernel setting lays in the Moore-Aronszajn
theorem \cite{Aronszajn1950}. It states that a Reproducing Kernel Hilbert
Space (RKHS) $\mathcal{H}$ can be associated to $\mathcal{X}$ and $k$ through
a mapping function $\phi$ from $\mathcal{X}$ to $\mathcal{H}$ such that
$\langle\phi(x),\phi(y)\rangle_{\mathcal{H}}=k(x,y)$ for all $x$ and $y$ in
$\mathcal{X}$. The mapping $\phi$ is called the \emph{feature map}. It enables
one to leverage the Hilbert structure of $\mathcal{H}$ in order to build
machine learning algorithms on $\mathcal{X}$ indirectly.This can be done in
general without using $\phi$ but rather by relying on $k$ only: this is known
as the \emph{kernel trick} (see
e.g. \cite{ShaweTaylorChristianini2004KernelMethods}).

Notice that the kernel can be used to define a dissimilarity on
$\mathcal{X}$ by transporting the Hilbert distance from
$\mathcal{H}$. Indeed, it is natural to define $d_k$ on $\mathcal{X}$ by
\begin{equation}
  \label{eq:distance:kernel}
d_k(x,y)=\langle\phi(x)-\phi(y),\phi(x)-\phi(y)\rangle_{\mathcal{H}}.
\end{equation}
Elementary algebraic manipulations show that 
\begin{equation}
  \label{eq:distance:kernel:trick}
d_k(x,y)=k(x,x)+k(y,y)-2k(x,y),
\end{equation}
which is an example of the use of the kernel trick to avoid using explicitly
$\phi$. 

The construction of $d_k$ shows that the dissimilarity setting is more general
than the kernel setting. It is always possible to use a kernel as the basis of
a dissimilarity: all the dissimilarity variants of the SOM can used on kernel
data. Therefore, we will focus mainly on dissimilarity algorithms, and then
discuss how they relate to their kernel counterparts. 

Notice finally that as in the case of the dissimilarity setting, the kernel
can be given as a function from $\mathcal{X}$ to $\mathbb{R}$ or as a kernel
matrix $K=(K_{ij})=(k(x_i,x_j))$. In the latter case, $K$ is symmetric and
positive definite and is associated to a dissimilarity matrix $D_K$ via
equation (\ref{eq:distance:kernel:trick}). 

\subsection{SOM}
To contrast its classical setting with the dissimilarity and kernel ones, and
to introduce our notations, we briefly recall the SOM principle and algorithm
\cite{KohonenSOM}. A SOM is a low dimensional clustered representation of a
data set.

One needs first to specify a low dimensional prior structure, in general a
regular lattice of $K$ units/neurons positioned in $\mathbb{R}^2$, the
$(r_k)_{1\leq k\leq K}$. The structure induces a time dependent neighborhood
function $h_{kl}(t)$ which measures how much the prototype/model associated to
unit $r_k$ should be close to the one associated to unit $r_l$, at step $t$ of the
learning algorithm (from 0 for unrelated models to 1 for maximally related
ones). We will not discuss here the numerous possible variants for this
neighborhood function \cite{KohonenSOM}: if the lattice is made of points
$r_k$ in $\mathbb{R}^2$ a classical choice is
\[
h_{kl}(t)=exp\left(-\frac{\|r_k-r_l\|^2}{2\sigma^2(t)}\right),
\]
where $\sigma$ increases over time to reduce gradually the influences of the neighbors
during learning. 

The SOM attaches to each unit/neuron $r_k$ in the prior structure a prototype/model
in the data space $m_k$. The objective of the SOM algorithm is to adapt the
values of the models in such a way that each data point is as close as
possible to its closest model in the data space (at standard goal in prototype
based clustering). In addition if the closest model for the data point
$x$ is $m_k$, then $m_l$ should also be close to $x$ if $r_k$ and $r_l$ are
close in the prior structure. In other words, proximities in the prior
structure should reflect proximities in the data space and vice versa. The
unit/neuron associated to the closest model of a data point is called the
\emph{best matching unit} (BMU) for this point. The set of points for which
$r_k$ is the BMU defines a cluster in the data space, denoted $C_k$. 

This is essentially achieved via two major algorithms (and dozens of
variants). Let us assume that the data space is a classical normed vector
space. Then both algorithms initialize the prototypes $(m_k)_{1\leq k\leq K}$
in an ``appropriate way'' and proceed then iteratively. We will not discuss
initialization strategies in this paper. 

In the stochastic/online SOM (SSOM), a data point $x$ is selected
randomly\footnote{or data points are looped through.} at each iteration
$t$. Then $c\in\{1,\ldots,K\}$ is determined as the index of the best matching
unit, that is
\begin{equation}
c=\arg\min_{k\in\{1,\ldots,K\}}\|x-m_k(t)\|^2,
\end{equation}
and all prototypes are updated via
\begin{equation}\label{eq:proto:sto}
m_k(t+1)=m_k(t)+\epsilon(t)h_{kc}(t)(x-m_k(t)),
\end{equation}
where $\epsilon(t)$ is a learning rate. 

In the batch SOM (BSOM), each iteration is made of two steps. In the first
step, the best matching unit for each data point $x_i$ is determined as:
\begin{equation}\label{eq:BMU:batch}
c_i(t)=\arg\min_{k\in\{1,\ldots,K\}}\|x_i-m_k(t)\|^2.
\end{equation}
Then all prototypes are updated via a weighted average
\begin{equation}\label{eq:proto:batch}
m_k(t+1)=\frac{\sum_{i=1}^Nh_{kc_i(t)}(t)x_i}{\sum_{i=1}^Nh_{kc_i(t)}(t)}.
\end{equation}
Obviously, neither algorithm can be applied \emph{as is} on non vector data. 

\section{The Median SOM}\label{sec:median-som}
\subsection{General principle}
It is well known (and obvious) that the prototype update step of the Batch SOM can
be considered as solving an optimization problem, namely
\begin{equation}\label{eq:optim:euclidean}
\forall\ k\in\{1,\ldots, K\},\ m_k(t+1)=\arg\min_{s}\sum_{i=1}^Nh_{kc_i(t)}(t)\|s-x_i\|^2.
\end{equation}
This turns the vector space operations involved in equation
(\ref{eq:proto:batch}) into an optimization problem that uses only the squared
Euclidean norm between prototypes and observations. In an arbitrary space
$\mathcal{X}$ with a dissimilarity, $\|s_k-x_i\|^2$ can be replaced by the
dissimilarity between $s_k$ and $x_i$ which turns problem
(\ref{eq:optim:euclidean}) into 
\begin{equation}
  \label{eq:optim:median}
\forall\ k\in\{1,\ldots, K\},\ m_k(t+1)=\arg\min_{s\in\mathcal{X}}\sum_{i=1}^Nh_{kc_i(t)}(t)d(s,x_i),
\end{equation}
which is a typical generalized median problem. 

However, the most general dissimilarity setting only assumes the availability
of dissimilarities between \emph{observations} not between arbitrary points in
$\mathcal{X}$. In fact, generating new points in $\mathcal{X}$ might be
difficult for complex data such as texts. Then the most general solution
consists in looking for the optimal prototypes into the data set rather than
in $\mathcal{X}$. The Median SOM
\cite{KohonenSymbol1996,KohonenSomervuo1998Symbol,KohonenSomervuo2002NonVectorial}
and its variants
\cite{ElGolliConanGuezRossi04JSDA,ElGolliConanGuezRossiIFCS2004SomDiss} are
based on this principle. The Median SOM
consists in iterating two steps. In the first step, the best matching unit for
each data point $x_i$ is determined as
\begin{equation}\label{eq:BMU:median}
c_i(t)=\arg\min_{k\in\{1,\ldots,K\}}d(x_i,m_k(t)).
\end{equation}
Then all prototypes are updated by solving the generalized median problem
\begin{equation}
  \label{eq:optim:median:constraint}
\forall\ k\in\{1,\ldots, K\},\ m_k(t+1)=\arg\min_{x_j}\sum_{i=1}^Nh_{kc_i(t)}(t)D_{ij}.
\end{equation}
Notice that each prototype is a data point which means that in equation
(\ref{eq:BMU:median}) $d(x_i,m_k(t))$ is in fact a $D_{il}$ for some $l$. 

A variant of the Median SOM was proposed in \cite{AmbroiseGovaert1996}: rather
than solving problem \eqref{eq:optim:median:constraint}, it associates to each
unit the generalized median of the corresponding cluster (in other words, it
does not take into account the neighborhood structure at this point). Then
the BMU of a data point is chosen randomly using the neighborhood structure
and the dissimilarities. This means that a data point can be moved from its
natural BMU to a nearby one. As far as we know, this variant has not been
studied in details. 

\subsection{Limitations of the Median SOM}\label{sec:limit-medi-som}
The Median SOM has numerous problems. As a batch SOM it is expected to request
more iterations to converge than a potential stochastic version (which is
not possible in the present context, unfortunately). In addition, it will also
exhibit sensitivity to its initial configuration. 

There are also problems more specific to the Median SOM. Each iteration of the
algorithm has a rather high computational cost: a naive implementation leads
to a cost of $O(N^2K+NK^2)$ per iteration, while a more careful one still
costs $O(N^2+NK^2)$ \cite{conan-guezrossietal2006neural-networks}. Numerous
tricks can be used to reduce the actual cost per
iteration \cite{conan-guezrossi2007speeding-up,conan-guezrossi2008nouvelles-technologies}
but the $N^2$ factor cannot be avoided without introducing approximations.

Arguably the two main drawbacks of the Median SOM are of a more intrinsic
nature. Firstly, restricting the prototypes to be chosen in the data set has
some very adverse effects. A basic yet important problem comes from collisions
in prototypes \cite{rossi2007model-collisions}: two different units can have
the same optimal solution according to equation
(\ref{eq:optim:median:constraint}). This corresponds to massive folding of the
two dimensional representation associated to the SOM and thus to a sub-optimal
data summary. In addition, equation (\ref{eq:BMU:median}) needs a tie breaking
rule which will in general increase the cost of BMU determination (see
\cite{KohonenSomervuo2002NonVectorial} for an example of such a rule). The
solution proposed in \cite{rossi2007model-collisions} can be used to avoid
those problems at a reasonable computational cost. 

A more subtle consequence of the restriction of prototypes to data points is
that no unit can remain empty, apart from collided prototypes. Indeed, the BMU
of a data point that is used as a prototype should be the unit of which it is
the prototype. This means that no interpolation effect can take place in the
Median SOM \cite{Somervuo2003SymbolAverage,Somervuo2004OnlineSymbol} a fact
that limits strongly the usefulness of visual representations such as the
U-matrix \cite{UltschSiemon1990Umatrix,ultsch2005esom}. For some specific data types such as
strings, this can be avoided by introducing ways of generating new data points
by some form of interpolations. This was studied in
\cite{Somervuo2003SymbolAverage,Somervuo2004OnlineSymbol} together with a
stochastic/online algorithm. 

A generic solution to lift the prototype restriction is provided by the relational SOM
described in Section \ref{sec:relational-som}. 

\subsection{Non metric dissimilarities}\label{sec:non-metr-diss}
The second intrinsic problem of the Median SOM is its reliance on
a prototype based representation of a cluster in the dissimilarity context,
while this is only justified in the Euclidean context. Indeed let us consider
that the $N$ data points $(x_i)_{1\leq i\leq N}$ belong to a Euclidean
space. Then for any vector of positive weights $(\beta_i)_{1\leq i\leq N}$,
the well known König-Huygens identity states:
\begin{equation}\label{eq:variance:distances}
  \sum_{i=1}^N\beta_i\left\|\frac{\sum_{j=1}^N\beta_jx_j}{\sum_{j=1}^N\beta_j}-x_i\right\|^2=
\frac{1}{2}\frac{1}{\sum_{i=1}^N\beta_i}\sum_{i=1}^N\sum_{j=1}^N
\beta_i\beta_j\|x_i-x_j\|^2.
\end{equation}
This means that
\begin{equation}
  \label{eq:som:inertia:proto}
\min_{m} \sum_{i=1}^N\beta_i\left\|m-x_i\right\|^2=
\frac{1}{2}\frac{1}{\sum_{i=1}^N\beta_i}\sum_{i=1}^N\sum_{j=1}^N
\beta_i\beta_j\|x_i-x_j\|^2.
\end{equation}
Applied to the SOM, this means that solving\footnote{The quantity optimized in
equation \eqref{eq:som:problem:proto} is the energy defined in \cite{Heskes93}.}
\begin{equation}
  \label{eq:som:problem:proto}
(m(t),c(t))=\arg\min_{m,c}\sum_{k=1}^K\sum_{i=1}^Nh_{kc_i}(t)\|m_k-x_i\|^2,
\end{equation}
where $m(t)=(m_1(t),\ldots,m_K(t))$ denotes the prototypes and
$c=(c_1,\ldots,c_n)$ denotes the BMU mapping, is equivalent to solving
\begin{equation}
  \label{eq:som:problem:inertia}
c(t)=\arg\min_{c}\frac{1}{2}\sum_{k=1}^K\frac{1}{\sum_{i=1}^Nh_{kc_i}(t)}\sum_{i=1}^N\sum_{j=1}^N
h_{kc_i}(t)h_{kc_j}(t)\|x_i-x_j\|^2.
\end{equation}
This second problem makes clear that the classical SOM is not only based on
\emph{quantization} but is also optimizing the within pairwise distances in
the clusters defined by the BMU mapping. Here $h_{kc_i}$ is considered as a
form of membership value of $x_i$ to cluster $k$, which give the ``size''
$\sum_{i=1}^Nh_{kc_i}$ to the cluster $k$. Then the sum of pairwise distances
in each cluster measures the compactness of the cluster in terms of within
variance. As the SOM minimizes the sum of those quantities, it can be seen as
a \emph{clustering} algorithm\footnote{This classical analysis mimics the one
  used to see the k-means algorithm both as a clustering algorithm and as a
  quantization algorithm.}.

However, the  König-Huygens identity does not apply to arbitrary
dissimilarities. In other words, the natural dissimilarity version of problem
\eqref{eq:som:problem:inertia} that is 
\begin{equation}
  \label{eq:som:problem:inertia:dissimilarity}
c(t)=\arg\min_{c}\frac{1}{2}\sum_{k=1}^K\frac{1}{\sum_{i=1}^Nh_{kc_i}(t)}\sum_{i=1}^N\sum_{j=1}^N
h_{kc_i}(t)h_{kc_j}(t)d(x_i,x_j),
\end{equation}
is not equivalent to the Median SOM problem given by 
\begin{equation}
  \label{eq:som:problem:proto:dissimilarity}
(m(t),c(t))=\arg\min_{m\in\{x_1,\ldots,x_N\}^K,c}\sum_{k=1}^K\sum_{i=1}^Nh_{kc_i}(t)d(x_i,m_k).
\end{equation}
When the dissimilarity satisfies the triangular inequality
this is not a major problem. By virtue of the triangular inequality, we
have for all $m$
\begin{equation}
  \label{eq:triangular}
 d(x_i,x_j)\leq d(x_i,m)+d(m,x_j),
\end{equation}
and therefore for all $m$
\begin{equation}
  \label{eq:triangular:som}
\sum_{i=1}^N\sum_{j=1}^N h_{kc_i}(t)h_{kc_j}(t)d(x_i,x_j)\leq 2\left(\sum_{i=1}^Nh_{kc_i}(t)\right)\sum_{j=1}^N h_{kc_j}(t)d(x_j,m),
\end{equation}
which shows that
\begin{equation}
  \label{eq:triangular:som:proxy}
\frac{1}{2\sum_{i=1}^Nh_{kc_i}(t)}  \sum_{i=1}^N\sum_{j=1}^N
h_{kc_i}(t)h_{kc_j}(t)d(x_i,x_j)\leq \min_{m}\sum_{j=1}^N h_{kc_j}(t)d(x_j,m).
\end{equation}
Then the Median SOM is optimizing an upper bound of the cluster oriented
quality criterion for dissimilarities. In practice, this means that a good
quantization will give compact clusters. 

However, when the dissimilarity does not satisfy the triangular inequality,
the two criteria are not directly related any more. In fact, one prototype can
be close to a set of data points while those points remain far apart from each
other. Then doing of form of \emph{quantization} by solving problem
\eqref{eq:som:problem:proto:dissimilarity} is not the same thing as doing a
form of \emph{clustering} by solving problem
\eqref{eq:som:problem:inertia:dissimilarity}. By choosing the prototype based
solution, the Median SOM appears to be a quantization method rather than a
clustering one. If the goal is to display \emph{prototypes} in an organized
way, then this choice make sense (but must be explicit). If the goal is to
display \emph{clusters} in an organized way, this choice is intrinsically
suboptimal. As pointed out in Section \ref{sec:discussion}, dissimilarity SOMs
are not very adapted to prototype display, which puts in question the interest
of the Median SOM in particular and of the quantization approach in general.

\section{The Relational SOM}\label{sec:relational-som}
The quantification of the prototypes induced by restricting them to data
points has quite negative effects described in Section
\ref{sec:limit-medi-som}. The relational approach is a way to address this
problem. It is based on the simple following observation
\cite{HathawayEtAl1989RelationCmeans}. Let the $(x_i)_{1,\ldots,N}$ be $N$
points in a Hilbert space equipped with the inner product $\langle.,.\rangle$
and let $y=\sum_{i=1}^N\alpha_ix_i$ for arbitrary real valued coefficients
$\alpha^T=(\alpha_i)_{1,\ldots,N}$ with $\sum_{i=1}^N\alpha_i=1$. Then
\begin{equation}
  \label{eq:relational}
\langle x_i-y,x_i-y\rangle=(D\alpha)_i-\frac{1}{2}\alpha^TD\alpha,
\end{equation}
where $D$ is the squared distance matrix given by $D_{ij}=\langle
x_i-x_j,x_i-x_j\rangle$. This means that computing the (squared) distance between a
linear combination of some data points and any of those data points can be
done using only the coefficients of the combination and the (squared) distance
matrix between those points.

\subsection{Principle}\label{sec:relational:principle}
But as shown by equation (\ref{eq:proto:batch}), prototypes in
the classical SOM are exactly linear combinations of data points whose
coefficients sum to one. It is therefore possible to express the Batch SOM
algorithm without using directly the values of the $x_i$, but rather by
keeping the coefficients of the prototypes and using equation
(\ref{eq:relational}) and the squared distance matrix to perform all
calculations.

Then one can simply apply the so called \emph{relational} version of the
algorithm to an arbitrary dissimilarity matrix as if it were a squared
euclidean one. This is essentially what is done in
\cite{HathawayBezdek1994NerfCMeans,HathawayEtAl1989RelationCmeans} for the
c-means (a fuzzy variant of the k-means) and in
\cite{hammerhasenfussetal2007topographic-processing} for the Batch SOM (and
the Batch Neural Gas \cite{cottrell2006batch}). Using the concept of pseudo-Euclidean spaces, it was
shown in \cite{HammerHasenfuss2010NeuralComputation} that this general
approach can be given a rigorous derivation: it amounts to using the original
algorithm (SOM, k-means, etc.) on a pseudo-Euclidean embedding of the data
points.

In practice, the Batch relational SOM proceeds by iterating two steps that are
very similar to the classical Batch SOM steps. The main difference is that
each prototype $m_k(t)$ (at iteration $t$) is given by a vector of
$\mathbb{R}^N$, $\alpha_k(t)$, which represents the coefficients of the linear
combination of the $x_i$ in the pseudo-Euclidean embedding. Then the best
matching unit computation from equation (\ref{eq:BMU:batch}) is replaced by
\begin{equation}
  \label{eq:BMU:batch:relational}
c_i(t)=\arg\min_{k\in\{1,\ldots,K\}}\left((D\alpha_k(t))_i-\frac{1}{2}\alpha_k(t)^TD\alpha_k(t)\right),
\end{equation}
while the prototype update becomes
\begin{equation}\label{eq:proto:batch:relational}
\alpha_k(t+1)_i=\frac{h_{kc_i(t)}}{\sum_{l=1}^Nh_{kc_l(t)}}.
\end{equation}
A stochastic/online variant of this algorithms was proposed in
\cite{OlteanuEtAl2013SRSOM}. As for the classical SOM, it consists in
selecting randomly a data point $x_i$, computing its BMU $c_i$ (using equation
(\ref{eq:BMU:batch:relational})) and updating all prototypes as follows:
\begin{equation}
  \label{eq:proto:sto:relational}
\alpha_k(t+1)_j=\alpha_k(t)_j+\epsilon(t)h_{kc_i}(t)(\delta_{ij}-\alpha_k(t)_j),
\end{equation}
where $\delta_{ij}$ equals 1 when $i=j$ and 0 in other cases. Notice that is
the $\alpha_k$ are initialized so as to sum to one, this is preserved by this
update. As shown in
\cite{OlteanuEtAl2013SRSOM}, the stochastic variant tends to be less sensitive
to the initial values of the prototypes. However \cite{OlteanuEtAl2013SRSOM}
overlooks that both batch and online relational SOM algorithms share the same
computational cost per iteration\footnote{the cost reported in
  \cite{OlteanuEtAl2013SRSOM} for the batch relational SOM is incorrect.}
which negates the traditional computational gain provided by online versions.

\subsection{Limitations of the Relational SOM}
The Relational SOM solves several problems of the Median SOM. In particular,
it is not subject to the quantization effect induced by constraining the
prototypes to be data points. As a consequence, it exhibits in practice the
same interpolation effects as the classical SOM. The availability of a
stochastic version provides also a simple way to reduce the adverse effects of
a bad initialization. 

However, the relational SOM is very computationally intensive. Indeed, the
evaluation of all the $\alpha_k(t)^TD\alpha_k(t)$ costs $O(KN^2)$
operations. Neither the dissimilarity matrix nor the prototype coefficients
are sparse and there is no way to reduce this costs without introducing
approximations. Notice that this cost is per iteration in both the batch and
the stochastic versions of the relational SOM. This is $K$ times larger than
the Median SOM. 

This large cost has motivated research on approximation techniques such as
\cite{rossihasenfussetal2007accelerating-relational}. The most principled
approach consists in approximating the calculation of the matrix product via
the Nyström technique \cite{williams2001using}, as explored in
\cite{HammerEtAlWSOM2011}.  

\section{Soft Topographic Mapping for Proximity
  Data}\label{sec:soft-topogr-mapp}
As pointed out in Section \ref{sec:non-metr-diss}, if an algorithm relies on
prototypes with a general possibly non metric dissimilarity, it provides only
quantization and not clustering.  When organized clusters are looked for, one
can try to solve problem \eqref{eq:som:problem:inertia:dissimilarity}
directly, that is without relying on prototypes. 

\subsection{A deterministic annealing scheme}
However problem \eqref{eq:som:problem:inertia:dissimilarity} is combinatorial
and highly non convex. In particular, the absence of prototypes rules out
standard alternating optimization schemes. Following the analysis done in the
case of the dissimilarity version of the k-means in
\cite{BuhmannHofmann1994ICPR,HofmannBuhmann1997TPAMI}, Graepel et
al. introduce in \cite{GraepelEtAl1998GSOM,GraepelObermayer1999DSOM} a
deterministic annealing approach to address problem
\eqref{eq:som:problem:inertia:dissimilarity}. The approach introduces a mean
field approximation which estimates by $e_{ik}$ the effects in the criterion
of problem \eqref{eq:som:problem:inertia:dissimilarity} of assigning the data
point $x_i$ in cluster $k$. In addition, it computes soft assignments to the
cluster/unit, denoted $\gamma_{ik}$ for the membership of $x_i$ to cluster $k$
($\gamma_{ik}\in[0,1]$ and $\sum_{k=1}^K\gamma_{ik}=1$).  The optimal mean
field is given by
\begin{equation}
  \label{eq:mean:field}
e_{ik}=\sum_{s=1}^Kh_{ks}\sum_{j=1}^Nb_{js}\left(d(x_i,x_j)-\frac{1}{2}\sum_{l=1}^Nb_{ls}d(x_j,x_l)\right),  
\end{equation}
where the $b_{js}$ are given by
\begin{equation}
  \label{eq:mean:field:b}
b_{js}=\frac{\sum_{k=1}^K\gamma_{jk}h_{ks}}{\sum_{i=1}^N\sum_{k=1}^K\gamma_{ik}h_{ks}}.  
\end{equation}
Soft assignments are updated according to 
\begin{equation}
  \label{eq:mean:field:soft}
\gamma_{ik}=\frac{\exp(-\beta e_{ik})}{\sum_{s=1}^K\exp(-\beta e_{is})},
\end{equation}
where $\beta$ is an annealing parameter. It plays the role of an inverse
temperature and is therefore gradually increased at each step of the
algorithm. 

In practice, the so-called Soft Topographic Mapping for Proximity Data (STMP)
is trained in an iterative batch like procedure. Given an annealing schedule
(that is a series of increasing values for $\beta$) and initial random values
of the mean field, the algorithm iterates evaluating equation
(\ref{eq:mean:field:soft}), then equation (\ref{eq:mean:field:b}) and finally
equation (\ref{eq:mean:field}) for a fixed value of $\beta$, until
convergence. When this convergence is reached, $\beta$ is increased and the
iterations restart from the current value of the mean field.

Notice in equation \eqref{eq:mean:field:b} that the neighborhood function is
\emph{fixed} in this approach, whereas it is evolving with time in most SOM
implementations.

\subsection{Limitations of the STMP}
It is well known that the quality of the results obtained by deterministic
annealing are highly dependent on the annealing scheme
\cite{RoseDeterministicAnnealing1999}. It is particularly important to avoid
missing transition phases. Graepel et al. have analyzed transition phases in
the STMP in \cite{GraepelObermayer1999DSOM}. As in
\cite{RoseDeterministicAnnealing1999,HofmannBuhmann1997TPAMI}, the first
critical temperature is related to a dominant eigenvalue of the dissimilarity
matrix. As this is in general a dense matrix, the minimal cost of computing the
critical temperature is $O(N^2)$. In addition, each internal iteration of the
algorithm is dominated by the update of the mean field according to equation
(\ref{eq:mean:field}). The cost of a full update is in $O(N^2K+NK^2)$. The STMP is
therefore computationally intensive. It should be noted however that an
approximation of the mean field update that reduces the cost to $O(N^2K)$ is proposed in
\cite{GraepelObermayer1999DSOM}, leading to the same computational cost as the
relational SOM.

In addition, as will appear clearly in Section \ref{sec:stmp-prototype-based},
the STMP is based on prototypes, even they appear only indirectly. Therefore
while it tries to optimize the clustering criterion associated to the SOM, it
resorts to a similar quantization quality proxy as the relational SOM. 

\section{Kernel SOM}\label{sec:kernel-som}
As recalled in Section \ref{sec:kernel-data}, the kernel setting is easier to
deal with than the dissimilarity one. Indeed the embedding into a Hilbert
space $\mathcal{H}$ enables to apply any classical machine learning
method to kernel data by leveraging the Euclidean structure of
$\mathcal{H}$. The kernel trick allows one to implement those methods
efficiently. 

\subsection{The kernel trick for the SOM}
In the case of the SOM, the kernel trick is based on the same fundamental remark
that enables the relational SOM (see Section \ref{sec:relational:principle}):
in the Batch SOM, the prototypes are linear combinations of the data
points. If the initial values of the prototypes are linear combinations of the data
points (and not random points), this is also the case for the
stochastic/online SOM. 

Then assume given a kernel $k$ on $\mathcal{X}$, with its associated Hilbert
space $\mathcal{H}$ and mapping $\phi$. Implementing the Batch SOM in
$\mathcal{H}$ means working on the mapped data set $(\phi(x_i))_{1\leq i\leq
  N}$ with prototypes $m_k(t)$ of the form
$m_k(t)=\sum_{i=1}^N\alpha_{ki}(t)\phi(x_i)$. Then equation (\ref{eq:BMU:batch})
becomes
\begin{equation}
\label{eq:BMU:batch:kernel}
c_i(t)=\arg\min_{k\in\{1,\ldots,K\}}\|\phi(x_i)-m_k(t)\|^2_{\mathcal{H}},
\end{equation}
with
\begin{align}
  \label{eq:dist:kernel:proto}
\|\phi(x_i)-m_k(t)\|^2_{\mathcal{H}}=& k(x_i,x_i)-2\sum_{j=1}^N\alpha_{kj}(t)k(x_k,x_j)\\
&+\sum_{j=1}^N\sum_{l=1}^N\alpha_{kj}(t)\alpha_{kl}(t)k(x_j,x_l).\notag
\end{align}
Equation (\ref{eq:dist:kernel:proto}) is a typical result of the kernel trick:
computing the distance between a data point and a linear combination of the
data points can be done using solely the kernel function (or matrix). To our
knowledge, the first use of the kernel trick in a SOM context was made in
\cite{GraepelEtAl1998GSOM}. 

Notice that equation (\ref{eq:proto:batch}) can also been implemented without
using explicitly the mapping $\phi$ as one needs only the coefficients of the
linear combination which are given by
\begin{equation}
  \label{eq:kernel:coefficients}
\alpha_{ki}(t+1)=\frac{h_{kc_i(t)}}{\sum_{l=1}^Nh_{kc_l(t)}},
\end{equation}
exactly as in equation (\ref{eq:proto:batch:relational}). While the earliest
kernel SOM (STMK) in \cite{GraepelEtAl1998GSOM} is optimized using deterministic
annealing (as the SMTP presented in Section \ref{sec:soft-topogr-mapp}), the
kernel trick enables the more traditional online SOM
\cite{macdonald_fyfe_ICKIESAT2000} and batch SOM
\cite{BouletEtAl2008Neurocomputing,MartinMerinoMunoz2004BatchKSOM,villarossi2007comparison-between}
derived from the previous equations.

It should be noted for the sake of completeness that another kernel SOM was
proposed in \cite{andras_IJNS2002}. However, this variant assumes that
$\mathcal{X}$ is a vector space and therefore is not applicable to the present
setting. 

\subsection{Limitations of the kernel SOM}
As it is built indirectly on a Hilbert space embedding, the kernel SOM does
not suffer from constrained prototypes. The stronger assumptions made on
kernels compared to dissimilarities guarantee the equivalence between finding
good prototypes and finding compact clusters. Kernel SOM has also both online
and batch versions. 

Then the main limitation of the kernel SOM is its computational cost. Indeed,
as for the relational SOM, evaluating the distances in equation
(\ref{eq:dist:kernel:proto}) has a $O(KN^2)$ cost. The approximation schemes
proposed for the relational SOM
\cite{HammerEtAlWSOM2011,rossihasenfussetal2007accelerating-relational} can be
used for the kernel SOM at the cost of reduced performances in terms of data
representation.

\section{Equivalences between SOM  variants}\label{sec:equiv-betw-som}
It might seem at first that all the variants presented in the previous
sections are quite different, both in terms of goals and algorithms. On the
contrary, with the exception of the Median SOM which is very specific in some
aspects, the variations between the different methods are explained by
optimization strategies rather than by hypothesis on the data.

\subsection{Relational and kernel methods are equivalent}
We have already pointed out that relational SOM and kernel SOM share the very
same principle of representing prototypes by a linear combination of the data
points. Both cases use the same coefficient update formulas whose structure
depends only on the type of the algorithm (batch or online).

The connections are even stronger in the sense that given a kernel, the
relational SOM algorithm obtained by using the dissimilarity associated to the
kernel is \emph{exactly} identical to the kernel SOM algorithm. Indeed if $K$
is the kernel matrix, then the dissimilarity matrix is given by
$D_{ij}=K_{ii}+K_{jj}-2K_{ij}$. Then for all $\alpha\in\mathbb{R}^N$ such that
$\sum_{i=1}^N\alpha_i=1$ and for all $i\in\{1,\ldots,N\}$
\begin{multline*}
(D\alpha)_i-\frac{1}{2}\alpha^TD\alpha=\sum_{j=1}^ND_{ij}\alpha_j-\frac{1}{2}\sum_{j=1}^N\sum_{l=1}^N\alpha_j\alpha_lD_{jl}\\
=\sum_{j=1}^N(K_{ii}+K_{jj}-2K_{ij})\alpha_j-\frac{1}{2}\sum_{j=1}^N\sum_{l=1}^N\alpha_j\alpha_l(K_{jj}+K_{ll}-2K_{jl})
\end{multline*}
Using $\sum_{i=1}^N\alpha_i=1$, the first term becomes
\begin{equation*}
\sum_{j=1}^N(K_{ii}+K_{jj}-2K_{ij})\alpha_j=K_{ii}+\sum_{j=1}^NK_{jj}\alpha_j-2\sum_{j=1}^NK_{ij}\alpha_j.
\end{equation*}
The same condition on $\alpha$ shows that
\begin{equation*}
\sum_{j=1}^N\sum_{l=1}^N\alpha_j\alpha_lK_{jj}=\sum_{j=1}^NK_{jj}\alpha_j,
\end{equation*}
and that
\begin{equation*}
\sum_{j=1}^N\sum_{l=1}^N\alpha_j\alpha_lK_{ll}=\sum_{l=1}^NK_{ll}\alpha_l.
\end{equation*}
Therefore
\begin{equation*}
\sum_{j=1}^N\sum_{l=1}^N\alpha_j\alpha_l(K_{jj}+K_{ll}-2K_{jl})=2\sum_{j=1}^NK_{jj}\alpha_j-2\sum_{j=1}^N\sum_{l=1}^N\alpha_j\alpha_lK_{jl}.
\end{equation*}
Combining those equations, we end up with
\begin{equation}
  \label{eq:equivalence}
(D\alpha)_i-\frac{1}{2}\alpha^TD\alpha=K_{ii}-2\sum_{j=1}^NK_{ij}\alpha_j+\sum_{j=1}^N\sum_{l=1}^N\alpha_j\alpha_lK_{jl}.
\end{equation}
The second part of this equation is exactly
$\|\phi(x_i)-m\|^2_{\mathcal{H}}$ when $m=\sum_{j=1}^N\alpha_j\phi(x_j)$ as
recalled in equation (\ref{eq:dist:kernel:proto}). Therefore, the best
matching unit determination in the relational SOM according to equation
(\ref{eq:BMU:batch:relational}) is exactly equivalent to the BMU determination
in the kernel SOM according to equation (\ref{eq:BMU:batch:kernel}). This
shows the equivalence between the two algorithms (in both batch and online
variants).

This equivalence shows that the batch relational SOM
from \cite{hammerhasenfussetal2007topographic-processing} is a rediscovery of the
batch kernel SOM from \cite{MartinMerinoMunoz2004BatchKSOM}, while the online
relational SOM from \cite{OlteanuEtAl2013SRSOM} is a rediscovery of the online
kernel SOM from \cite{macdonald_fyfe_ICKIESAT2000}. Results from
\cite{HammerHasenfuss2010NeuralComputation} show that those rediscoveries are
in fact \emph{generalizations} of kernel SOM variants as they extend the
Hilbert embedding to the more general pseudo-Euclidean embedding. In practice,
there is no reason to distinguish the kernel SOM from the relational SOM. 

\subsection{STMP is a prototype based approach}\label{sec:stmp-prototype-based}
On the surface, the STMP described in Section \ref{sec:soft-topogr-mapp} looks
very different from relational/kernel approaches as it tries to address the
combinatorial optimization problem
\eqref{eq:som:problem:inertia:dissimilarity} rather than the different problem
\eqref{eq:som:problem:proto:dissimilarity} 
associated to the generalized median. However, as analyzed in details in
\cite{HammerHasenfuss2010NeuralComputation}, the STMP differs from the
relational approach only by the use of deterministic annealing, not by the
absence of prototypes.

A careful analysis of equations (\ref{eq:mean:field}) and
(\ref{eq:proto:batch:relational}) clarifies this point. Indeed, let us
consider $\alpha_s=(b_{js})_{1\leq j\leq N}^T$ as the coefficient vector for a
linear combination of the data points $x_j$ embedded in the pseudo-Euclidean
space associated to the dissimilarity matrix $D$. Then
\begin{equation*}
\sum_{j=1}^Nb_{js}\left(d(x_i,x_j)-\frac{1}{2}\sum_{l=1}^Nb_{ls}d(x_j,x_l)\right)=(D\alpha_s)_i-\frac{1}{2}\alpha_s^TD\alpha_s.
\end{equation*}
The right hand part is the distance in the pseudo-Euclidean space between the
prototype associated to $\alpha_s$ and $x_i$. Then $e_{ik}$ in equation
\eqref{eq:mean:field} is a weighted
average of distances between $x_i$ and each of the $\alpha_s$, where the
weights are given by the neighborhood function. As pointed out in
\cite{HammerHasenfuss2010NeuralComputation}, this can be seen as a relational
extension of the assignment rule proposed by Heskes and Kappen in
\cite{Heskes93}. 

However, rather than using crisp assignments to a best matching unit with the
lowest value of $e_{ik}$, the STMP uses a soft maximum strategy implemented by
equation (\ref{eq:mean:field:soft}) to obtain assignment probabilities
$\gamma_{ik}$. Those are used in turn to update the coefficients of the
prototypes in equation (\ref{eq:mean:field:b}). 

In fact the three algorithms proposed in \cite{GraepelEtAl1998GSOM} are all
based on the same deterministic annealing scheme, with an initial
implementation in $\mathbb{R}^p$ (the STVQ) and two generalization in the Hilbert space
associated to a kernel (STMK) and in the pseudo-Euclidean space associated to a
dissimilarity (STMP). The discussion of the previous section shows that the kernel
and the dissimilarity variants are strictly equivalent. 

\subsection{Summary}
We summarizes in the following tables the variants of the SOM discussed in this
paper. Table \ref{tab:som:variants} maps a data type and an optimization
strategy to a SOM variant. Relational variants include here the kernel
case. Table \ref{tab:som:variants:cost} gives the computational costs
of one iteration of the SOM variants. 

\setlength{\tabcolsep}{7pt}
\begin{table}[htbp]
  \begin{center}
    \begin{tabular}{llccc}\toprule
      &&\multicolumn{3}{c}{Data type}\\
      &&$\mathbb{R}^p$ data & Kernel & Dissimilarity\\\cmidrule(r){3-5}
      \multirow{3}{6em}{Optimization strategy}& Online & online SOM &
      \multicolumn{2}{c}{online relational SOM \cite{macdonald_fyfe_ICKIESAT2000,OlteanuEtAl2013SRSOM}}\\
      & Batch & batch SOM & \multicolumn{2}{c}{batch relational SOM \cite{hammerhasenfussetal2007topographic-processing,MartinMerinoMunoz2004BatchKSOM}}\\
      & Batch & NA & NA & Median SOM \cite{KohonenSymbol1996}\\
      & \multicolumn{1}{p{6em}}{Deterministic annealing}& STVQ
      \cite{GraepelEtAl1998GSOM} & STMK \cite{GraepelEtAl1998GSOM} & STMP \cite{GraepelEtAl1998GSOM}\\
      \bottomrule
    \end{tabular}
  \end{center}
  \caption{Variants of the SOM}
  \label{tab:som:variants}
\end{table}

\begin{table}[htbp]
  \begin{center}
    \begin{tabular}{lcc}\toprule
Algorithm & Assignment cost & Prototype update cost\\
\midrule
Batch SOM & $O(NKp)$ & $O(NKp)$ \\
Online SOM & $O(Kp)$ & $O(Kp)$ \\
Median SOM & $O(NK)$ & $O(N^2+NK^2)$\\
Batch relational SOM & $O(N^2K)$ & $O(NK)$\\
Online relational SOM & $O(N^2K)$ & $O(NK)$\\
STVQ &$O(NKp+NK^2)$&$O(NKp+NK^2)$\\
STMK/STMP &$O(N^2K+NK^2)$&$O(NK^2)$\\
\bottomrule
    \end{tabular}
  \end{center}
  \caption{Computational complexity of SOM variants for $N$ data points, $K$
    units and in $\mathbb{R}^p$ for the classical SOM.}
  \label{tab:som:variants:cost}
\end{table}

\section{Discussion}\label{sec:discussion}
Even if the kernel approaches are special cases of the relational ones, we
have numerous candidates for dissimilarity processing with the SOM. We discuss
those variants in this section.

\subsection{Median SOM}
In our opinion, there is almost no reason to use the Median SOM in practice,
except possibly the reduced computational burden compared to the relational
SOM ($O(N^2)$ compared to $O(N^2K)$ for the dominating terms). Indeed, the
Median SOM suffers from constraining the prototypes to be data points and
gives in general lower performances than the relational/kernel SOM as compared
to a ground truth or based on the usability of the results (see for
instance \cite{HammerEtAlWSOM2011,OlteanuEtAl2013SRSOM,villarossi2007comparison-between}). The
lack of interpolation capability is particularly damaging as it prevents in
general to display gaps between natural clusters with u-matrix like visual
representation \cite{UltschSiemon1990Umatrix,ultsch2005esom}. 

For large data sets, the factor $K$ increase in the cost of one iteration of
the relational SOM compared to the median SOM could be seen as a strong
argument for the latter. In our opinion, approximation techniques
\cite{HammerEtAlWSOM2011,rossihasenfussetal2007accelerating-relational} are
probably a better choice. This remains however to be tested as to our
knowledge the effects of the Nyström approximation have only been studied
extensively for the relational neural gas and the relational
GTM \cite{GisbrechtEtAlLinear2012,HammerEtAlWSOM2011,SchleifGisbrecht2013linear}.  

\subsection{Optimization strategy}
To our knowledge, no systematic study of the influence of the optimization
strategy has been conducted for SOM variants, even in the case of numerical
data. In this latter case, it is well known that the online/stochastic SOM is
less sensitive to initial conditions than the batch SOM. It is also generally
faster to converge and leads in general to a better overall topology
preservation \cite{fort2002advantages}. Similar results are observed in the
dissimilarity case in \cite{OlteanuEtAl2013SRSOM}. It should be noted however
that both analyses use only random initializations while it is well known (see
e.g. \cite{KohonenSOM}) that a PCA\footnote{PCA initialization is easily
  adapted to the relational case, as it was for kernel data
  \cite{scholkopf1997kernel}.} based initialization gives much better results
than a random one in the case of the batch SOM. It is also pointed in
\cite{KohonenSOM} that the neighborhood annealing schedule as some strong
effects on topology preservation in the batch SOM. Therefore, in terms of the
final quality of the SOM, it is not completely obvious that an online solution
will provide better results than a batch one.

In addition, the relational setting negates the computational advantage of the
online SOM versus the batch SOM. Indeed in the numerical case, one epoch of
the online SOM (a full presentation of all the data points) has roughly the
same cost as one iteration of the batch SOM. As the online SOM converges
generally with a very small number of epochs, its complete computational cost
is lower than the batch SOM. On the contrary, the cost of the relational SOM
is dominated by the calculation of $\alpha^TD\alpha$ in equation
\eqref{eq:BMU:batch:relational}. In the batch relational SOM this quantity can
be computed one time per prototype and per iteration, leading to a cost of
$O(N^2K)$ per iteration (this is overlooked in \cite{OlteanuEtAl2013SRSOM}
which reports erroneously a complexity of $O(N^3K)$ per iteration). In the
online version, it has also to be computed for each data point (because of the
prototype update that takes place after each data point presentation). This
means that one epoch of the online relational SOM costs $N$ times more than one
iteration of the batch relational SOM. We think therefore that a careful
implementation of the batch relational SOM should outperform the online
version, provided the initialization is conducted properly. 

Comparisons of the online/batch variants with the deterministic annealing
variants is missing, as far as we know. The extensive simulations conducted in
\cite{HammerHasenfuss2010NeuralComputation} compare the relational neural gas
to the dissimilarity deterministic annealing clustering of
\cite{BuhmannHofmann1994ICPR,HofmannBuhmann1997TPAMI}. Their conclusion is the
one expected from similar comparisons done on numerical data
\cite{RoseDeterministicAnnealing1999}: the sophisticated annealing strategy of
deterministic annealing techniques leads in general to better solutions
provided the critical temperatures are properly identified. This comes with a
largely increased cost, not really because of the cost per iterations but
rather because the algorithm comprises two loops: an inner loop for a given
temperature and an outer annealing loop. Therefore the total number of
iterations is in general of an order of magnitude higher than with classical
batch algorithms (see also \cite{rossivilla-vialaneix2010neurocomputing} for
similar results in the context of a graph specific variant of the SOM
principle). It should be also noted that in all deterministic variants
proposed in \cite{GraepelEtAl1998GSOM}, the neighborhood function is not
adapted during learning. The effects of this choice on the usability of the
final results remain to be studied.

To summarize, our opinion is that one should prefer a careful implementation
of the batch relational SOM, paired with a PCA like algorithm for
initialization and using the Nyström approximation for large data
sets. Further experimental work is needed to validate this choice.

\subsection{Clustering versus quantization}
As explained in Section \ref{sec:non-metr-diss}, an algorithm that resorts
(directly or indirectly) on prototypes for an arbitrary dissimilarity does in
fact of form of \emph{quantization} rather than a form of \emph{clustering}. To our
knowledge, no attempt has been made to minimize directly the prototype free
criterion used in problem \eqref{eq:som:problem:inertia:dissimilarity} and we
can only speculate on this point.

We should first note that in the case of classical clustering, it has been
shown in \cite{conan-guezrossi2012dissimilarity-clustering} that optimizing
directly the criterion from problem
\eqref{eq:som:problem:inertia:dissimilarity} in its k-means simplified form
gives better results than using the relational version of the k-means. While
the computational burden of both approaches are comparable, the direct
optimization of the pairwise dissimilarities criterion is based on a much
more sophisticated algorithm which combines state-of-the-art hierarchical
clustering \cite{Mullner11ModernHierarchical} with multi-level refinement from
graph clustering \cite{Hendrickson95MultilevelAlgorithm}.

Assuming such a complex technique could be used to train a SOM like algorithm,
one would obtain in the end a set of non empty clusters, organized according
to a lattice in 2 dimensions, something similar to what can be obtained with
the Median SOM. While the clusters would have a better quality, no
interpolation between them would be possible, as in the Median SOM. 

\subsection{How useful are the results?}
In our personal opinion, the main interest of the SOM is to provide rich
and yet readable visual representations of complex data
\cite{Vesanto1999SomVisu,VesantoPhD2002}. Unfortunately, the visualization
possibilities are reduced in the case of dissimilarity data. 

The main limitation is that for arbitrary data in an abstract space
$\mathcal{X}$, one cannot assume that an element of $\mathcal{X}$ can be
easily represented visually. Then even the Median SOM prototypes (which are
data points) cannot be visualized. As the prototypes (in all the variants)
do not have meaningful coordinates, component planes cannot be used. 

In fact, the only aspects of the results that can be displayed as in the case
of numerical data are dissimilarities between prototypes (in U matrix like
displays \cite{UltschSiemon1990Umatrix}) as well as numerical characteristics
of the clusters (size, compactness, etc.). But as pointed out in
\cite{ultsch2005esom}, among others, this type of visualization is interesting
mainly when the SOM uses a large number of units. While this is possible with
the relational SOM, it implies a high computational because of the dominating
$O(N^2K)$ term. The case of deterministic annealing versions of the SOM is
even more problematic with the $O(NK^2)$ complexity term induced by the soft
memberships.

In some situations, specific data visualization techniques can be built upon
the SOM's results. For instance by clustering graph nodes via a
kernel/dissimilarity SOM, one can draw a clustered graph representation, as
was proposed in \cite{BouletEtAl2008Neurocomputing}. However, it has been
shown in this case that specialized models derived from the SOM
\cite{rossivilla-vialaneix2010neurocomputing} or simpler dual approaches based
on graph clustering and graph visualization
\cite{rossivilla-vialaneix2011societe-fran-caise} give in general better final
results. 

To summarize, our opinion is that the appeal of a generic dissimilarity SOM is
somewhat reduced by the limited visualization opportunity it offers, compared
to the traditional SOM. Further work is needed to explore whether classical
visualization techniques, e.g. brushing and linking
\cite{BeckerCleveland1987Brushing} could be used to provide more interesting
displays based on the dissimilarity SOM.

\section{Conclusion}
We have reviewed in this paper the main variants of the SOM that are adapted
to dissimilarity data and to kernel data. Following
\cite{HammerHasenfuss2010NeuralComputation}, we have shown that the variants
differ more in terms of their optimisation strategy that in other aspects. We
have recalled in particular that kernel variants are strictly identical to
their relational counterpart. Taking into account computational aspects and
known experimental results, our opinion is that the best solution is the batch
relational SOM coupled with a structured initialization (PCA like) and with
the Nyström approximation for large data sets and thus that we need one
dissimilarity/kernel SOM variant only. 

However, as discussed above, the practical usefulness of the dissimilarity SOM
is reduced compared to the numerical SOM as most of the rich visual
representations associated to the SOM of not available for its dissimilarity
version. Without improvement in its visual outputs, it is not completely clear
if the dissimilarity SOM serves a real practical purpose beyond its elegant
generality and simplicity.

\bibliography{biblio}
\bibliographystyle{splncs03}

\end{document}